\documentclass[11pt]{article}
\usepackage{acl2012}
\usepackage{times}
\usepackage{latexsym}
\usepackage{amsmath}
\usepackage{multirow}
\usepackage{url}
\usepackage{url}
\usepackage{array}
\usepackage{graphicx}
\usepackage{subfig}
\usepackage{arydshln}
\usepackage{color}
\usepackage{textcomp}

\newcommand{\com}[1]{}
\newcommand{\old}[1]{}
\newcommand{\camready}[1]{#1}
\renewcommand{\camready}[1]{}

\newcommand{\my}[1]{}

\newcommand{\isection}[2]{\section{#1}\label{ssec:#2}}

\newcommand{\secref}[1]{\textsection{~\ref{ssec:#1}}}

\begin{document}


\title {Separated by an Un-common Language: Towards Judgment Language Informed Vector Space Modeling}


\author{Ira Leviant \\
IE\&M faculty\\
Technion - IIT \\
{\tt ira.leviant@campus.technion.ac.il} \\\And
Roi Reichart \\
IE\&M faculty\\
Technion - IIT \\
{\tt roiri@ie.technion.ac.il} \\
}

\date{}

\date{}

\maketitle

\begin{abstract}

A common evaluation practice in the vector space models ({\sc vsm}s) literature is to measure the models' ability 
to predict human judgments about lexical semantic relations between word pairs. 
Most existing evaluation sets, however, consist of scores collected for English word pairs only, 
ignoring the potential impact of the \textit {judgment language} in which word pairs are presented 
on the human scores.

In this paper we translate two prominent evaluation sets, wordsim353 (association) and SimLex999 (similarity), 
from English to Italian, German and Russian and collect scores for each dataset from crowdworkers fluent 
in its language. Our analysis reveals that human judgments are strongly impacted by 
the judgment language. Moreover, we show that the predictions of monolingual {\sc vsm}s 
do not necessarily best correlate with human judgments made with the language  
used for model training, suggesting that models and humans are affected 
differently by the language they use when making semantic judgments. 
Finally, we show that in a large number of setups, 
multilingual {\sc vsm} combination results in improved correlations with human 
judgments, suggesting that multilingualism may partially compensate 
for the judgment language effect on human judgments.\footnote{All the datasets and related documents
produced in this work will be released upon acceptance of the paper.}


\end{abstract}

\isection {Introduction}{sec:intro}

\my{TODO: 1. experiments with larger corpora. 2. Can we explain why English is the 
best training language ?
3. "It would be interesting to take a closer look at the type of word pairs that were
robust across the different languages." 4. Can we say something about the differences between BOW and W2V 
(where each of them succeeds/fails) ? 5. Explain CCA in more detail (see response to EMNLP-rev1. 
more reviewers asked about this). 6. Most judges speak English as well. Can this bias the results ? 
7. Add 2015 citations. 8. There are details (some in footnotes some in text) we will have to verify (the footnote 
verification style you sent me does not address). 9. minor: 1. cite Roy's CoNLL paper.  2. Our judges are likely to be English speakers 3. Replace "multilingual training" with "combination"}
In recent years, there has been an immense interest in the 
development of {\it Vector Space Models ({\sc vsm}s)} for word meaning representation. 
Most {\sc vsm}s are based on the distributional hypothesis \cite{Harris:1954}, 
stating that words that occur in similar contexts tend to have similar meanings. 

{\sc vsm}s produce a vector representation for each word in the lexicon. 
A common evaluation practice for such models is to compute a score for each member of a word pair set
by applying a similarity function to the vectors of the words participating in the pair. 
The resulting score should reflect the degree to which one or more lexical relations between 
the words in the pair hold. The correlation between the model's scores and the scores generated 
by human evaluators is then computed. 

Humans as well as {\sc vsm}s may consider various languages when making their judgments and predictions. 
Recent research on multilingual approaches to {\sc vsm}s aims to exploit multilingual training 
(training with corpora written in different languages) to improve {\sc vsm} predictions. 
The resulting models are evaluated either against human scores, most often produced for word pairs presented 
to the human evaluators in English, or on multilingual text mining tasks (\secref{sec:previous}).
While works of the latter group do recognize the connection 
between the {\sc vsm} {\it training language ({\sc tl})} and the 
task's language, to the best of our knowledge no previous work systematically 
explored the impact of the {\it judgment language ({\sc jl})}, the language in which 
word pairs are presented to human evaluators, on human semantic judgments and on their correlation 
with {\sc vsm} predictions.

In this paper we therefore explore two open issues: (a) the effect of the {\sc jl} 
on the human judgment of semantic relations between words; and (b) 
the effect of the {\sc tl}(s) on the capability of {\sc vsm}s to predict human judgments 
generated with different {\sc jl}s.
To address these issues we translate two prominent datasets of English word pairs scored 
for semantic relations: WordSim353 ({\sc ws353}, \cite{Finkelstein:2001}), consisting of 353 word pairs scored 
for association, and SimLex999 ({\sc sl999}, \cite{Hill:2014}), consisting of 999 
word pairs scored for similarity.
For each dataset, the word pairs and the annotation guidelines are translated 
to three languages from different branches of the Indo-European 
language family: German (Germanic), Italian (Romance) and Russian (Slavic). 
We then employ the CrowdFlower crowdsourcing service \footnote {http://www.crowdflower.com/} to collect  
judgments for each set from human evaluators fluent in its {\sc jl} (\secref{sec:data}).

In \secref{sec:judgment-language} we explore the hypothesis that due to a variety of 
factors -- linguistic, cultural  and others -- the {\sc jl} should affect 
human generated association and similarity scores. 
Indeed, our results show that inter evaluator agreement is significantly higher
within a {\sc jl} than it is across {\sc jl}s.
This suggests that word association and similarity are {\sc jl} dependent. 

We then investigate (\secref{sec:training-language}) the connection between the {\sc vsm} {\sc tl} and 
the human {\sc jl}. We experiment with two
{\sc vsm}s that capture distributional co-occurrence statistics in different ways: 
a bag-of-word ({\sc bow}) model that is based on direct counts 
and the neural network (NN) based word2vec ({\sc w2v}, \cite{Mikolov:2013a}).
We train these models on monolingual comparable corpora from our four 
{\sc jl}s (\secref{sec:vsm-models}) and compare their predicted 
scores with the human scores produced for the various {\sc jl}s. 
Our analysis reveals fundamental differences between 
word association and similarity. For example, while for association the predictions of a {\sc vsm} trained 
on a given language best correlate with human judgments made with that language, for similarity 
some {\sc jl}s better correlate with all monolingual models than others.


Finally (\secref{sec:training-language-combination}), we explore how multilingual model combination affects 
the ability of {\sc vsm}s to predict human judgments for varios {\sc jl}s. Our results show 
a positive effect for a large number of {\sc tl} and {\sc jl} combinations, suggesting that 
multilingualism may partially compensate for the judgment language effect on human semantic judgments.

\isection{Previous Work}{sec:previous}

{\bf Vector Space Models and Their Evaluation.}  
Earlier {\sc vsm} work (see \cite{turney2010frequency}) designed word representations based on word co-location 
counts, potentially post-processed using 
techniques such as Positive Pointwise Mutual Information 
(PPMI) and dimensionality reduction methods.  
Recently, much of the focus has drifted to the development 
of NNs for representation learning 
\cite [inter alia]{Bengio:2003,Collobert:2008,Collobert:2011,huang2012improving,Mikolov:2013a,Mikolov:2013b,Levy:2014a,Pennington:2014}.

{\sc vsm}s have been evaluated in two main forms: 
(a) comparing model-based word pair scores with human judgments of various semantic relations. 
The model scores are generated by applying a similarity function, usually the cosine metric, 
to the vectors generated by the model for the words in the pair 
\cite[inter alia]{huang2012improving,Baroni:2014,Levy:2014a,Pennington:2014,Schwartz:2015}; 
and (b) evaluating the contribution of the generated vectors to NLP applications 
\cite[inter alia]{Collobert:2008,Collobert:2011,Pennington:2014}.

Several evaluation sets consisting of English word pairs scored by humans for semantic relations 
(mostly association and similarity) are in use for {\sc vsm} evaluation. Among these are 
{\sc rg-65} \cite{RG65}, {\sc mc-30} \cite{MC30}, {\sc ws353} (\cite{Finkelstein:2001}), 
{\sc yp-130} \cite{yang2006verb}, and {\sc sl999} \cite{Hill:2014}. 
Recently a few evaluation sets consisting of scored word pairs in languages other than English 
(e.g. Arabic, French, Farsi, German, Portuguese, Romanian and Spanish) were presented 
\cite[inter alia]{gurevych2005,zesch2006,hassan2009,schmidt2011,camacho2015,koper2015}. 
Most of these datasets, however, are translations of the English sets, where 
the original human scores produced for the original English set are kept. 
Even for those cases where evaluation sets were re-scored (e.g. \cite{camacho2015,koper2015})
our investigation of the {\sc jl} effect is much more thorough.\footnote{Although 
{\sc ws353} was translated to German and then scored with the German {\sc jl} \cite{koper2015}, we 
translated and scored the dataset again in order to keep the same translation and scoring decisions 
across our datasets. We applied the same considerations when re-scoring the original English versions 
of {\sc ws353} and {\sc sl999}.} A comprehensive list of these datasets, as well as of evaluation 
sets for word relations beyond word pair similarity and association 
\cite[inter alia]{mitchell-lapata:2008,bruni:2012,Baroni:2012},
is given at http://wordvectors.org/suite.php.


{\bf Multilingual VS Modeling.} Recently, there has been a growing interest in 
multilingual vector space modeling 
\cite[inter alia]{klementiev:2012,lauly2013learning,khapramultilingual,hermann2014multilingual,Hermann:2014:ICLR,Kocisky:2014,lauly2014autoencoder,al2014polyglot,faruqui2014improving,Coulmance:2015}.
These works train {\sc vsm}s on multilingual data, either parallel or not, or combine {\sc vsm}s trained on 
monolingual data.  The resulting models are evaluated either against human scores, most often
produced for word pairs presented to the human evaluators in English, or on multilingual text mining
tasks.

\isection{Multilingual Human Judgment Data}{sec:data}

Here we describe the data collection process, consisting of dataset
translation (\ref{sec:data-translation}) and scoring (\ref{sec:rescoring}). 
Our working datasets are {\sc ws353} \cite{Finkelstein:2001}
and {\sc sl999} \cite{Hill:2014}.\footnote{The original datasets and annotation guidelines are available at 
{\tiny http://www.cs.technion.ac.il/$\sim$gabr/resources/data/wordsim353/wordsim353.html}
and {\tiny http://www.cl.cam.ac.uk/~fh295/simlex.html} respectively.}


\subsection {Evaluation Sets Translation}
\label{sec:data-translation}

We started by translating the {\sc ws353} and {\sc sl999} scoring guidelines to the target languages. 
For each language the translation was done by two native speakers, 
and disagreements were solved through a discussion mediated by an experiment manager. 
An external evaluator, fluent in both the target language and in English then verified 
the translation quality.

The word pair translation process was more complicated. 
We followed the same protocol outlined above and further set a number of rules 
that guided our translators in challenging cases. 
Below we discuss the different types of translation ambiguities addressed 
in our guidelines.

{\bf Gender.} In some cases English does not make gender distinctions that some of the other languages do. 
For example, the English word \textit{cat} refers to both the female and the male cat while 
in Russian and Italian each gender has its own word (e.g. \textit{gatto} and \textit{gatta} in Italian). 
In such cases, if the other word in the English pair 
has a clear gender interpretation we followed this gender in the translation of both words, 
otherwise we chose one of the genders randomly and kept it fixed across the target languages.\footnote{We 
did not observe any case of gender disagreement between languages.} 

{\bf Word Senses.} It is common that some words in a given language have a sense set that 
is not conveyed by any of the words of another given language. 
For example, the English word \textit{plane}, from the {\sc ws353} pair 
\textit{(car,plane)}, has both the \textit{airplane} and the \textit{geometric plane} senses. 
However, to the best of our translators' knowledge, no German, Italian or Russian 
word has these two senses. 

We assume that when the authors of the evaluation sets paired two words, they referred to 
their closest senses. Therefore, like for gender, we used the other word in the pair for sense disambiguation. 
In our example, \textit{plane} is translated to the target language word 
which has the \textit{airplane} meaning (e.g. \textit{Flugzeug} in German, \textit{aeroplano} in Italian), 
since this sense  is closer to the meaning of \textit{car}.
%

In cases where the other word in the pair does not clearly disambiguate the sense of its polysemous counterpart, 
we randomly chose one of the latter word's senses, and kept it fixed across the target languages. 
Consider, for example, the {\sc sl999} pair \textit{(portray,decide)}. 
\textit{Portray} has three senses \footnote {\scriptsize{http://www.merriam-webster.com/dictionary/portray}} - 
one related to \textit{describing someone or something}, one related to 
\textit{showing in painting} and one to \textit{playing a character in a tv show, play or a movie}.
Since it was not clear to our translators how the word \textit{decide} can facilitate 
sense selection, we randomly chose the first sense and used it across target languages.



Sense disambiguation is done on a POS basis as well.
For example, in the pair \textit{(attempt,peace)} \textit{attempt} can be a verb or a noun, 
but none of these senses is necessarily closer to the 
meaning of \textit{peace}. 
In such cases, reasoning that words with the same POS tend to have a closer meaning, we used the 
interpretation of the polysemous word which has the same POS as the other word in the pair. 
That is, in the current example \textit{attempt} was assigned its noun sense, 
as \textit{peace} is a noun. 
Naturally, the target language translation 
of a given English word may also have multiple senses, some of which are not expressed by the English word. 
We guided our translators to avoid such translations whenever possible, 
although in a few cases that was impossible. 

{\bf Pair Exclusion.} We excluded some of the pairs from the evaluation sets in our experiments. 
Three pairs were excluded from {\sc ws353} due to translation difficulties.
The pairs \textit{(noon,midday)} and \textit{(coast, shore)} were excluded because 
none of the target languages includes two different words that convey the meaning of either set.
The pair \textit{(football,soccer)} was also excluded since it reflects a cultural distinction 
that is not made in the target languages. The resulting datasets in all four languages 
therefore consist of 350 word pairs.

For {\sc sl999} all 999 pairs were translated, scored and employed in the {\sc jl} analysis of 
\secref{sec:judgment-language}. However, as for 23 of the pairs at least one of the participating words 
did not appear in at least one of the {\sc vsm} training corpora (see \secref {sec:vsm-models}), 
we excluded these pairs from the analysis of the relations between the {\sc tl}s and 
the {\sc jl}s (\secref{sec:training-language} and \secref{sec:training-language-combination}).

{\bf Inter Translator Agreement.} 
The disagreement rates between our two translators for {\sc ws353} (700 words) and {\sc sl999} (1998 words) 
are (left parentheses for {\sc ws353}, right for {\sc sl999}):
Russian ((85 words, 12.1\%), (353 words, 17.7\%)), 
Italian ((57 words, 8.1\%), (196 words, 9.8\%)) and German ((113 words, 16.1\%), (396 words, 19.8\%)). 
To resolve disagreements, for each language we asked one of the translators to choose 
the translation which is more similar in meaning to the other word in the pair. 
If this is not possible, the translator was asked to choose the word which seems to her more common in the target language.



\subsection{Word Pair Scoring}
\label{sec:rescoring}

We next describe the word pair scoring process. In order to keep our analysis unbiased across {\sc jl}s, 
we scored {\sc ws353} and {\sc sl999} in all four languages, including English.
We divided each dataset to non-overlapping batches of 50 word pairs each (7 for {\sc ws353}, 
20 for {\sc sl999}, with one {\sc sl999} batch consisting of 49 pairs) 
and employed the crowdflower crowdsourcing service to recruit fluent speakers of each target language 
to score each batch. Evaluators were presented with the scoring guidelines translated 
to their {\sc jl} and were asked to score the pairs on a 0-10 scale.

We verified the quality of our evaluators through a three step process. First, for each 
{\sc jl} we only recruited evaluators who were located at a country where 
this language is the mother tongue of the majority of the population (i.e. US, Germany, Italy or Russia). 
Second, in order to make sure that our evaluators understand the task properly, 
we generated 7 tests for each language, each consisting of two word pairs that 
do not appear in the evaluation set. 
The participating pairs consisted of words that were either very similar or very 
dissimilar. Before scoring a batch of word pairs, each evaluator was presented with a randomly sampled test in its 
language and was asked to score its word pairs. 
Every evaluator that assigned a similar pair with a score lower than 7 or a non-similar 
pair with a score higher than 3 was excluded from the experiment. 
Finally, we ran an outlier detection procedure in order to exclude evaluators 
whose scores were substantially different from those of the other evaluators of their batch.
\footnote {Some works that employ crowdsourcing compare some of the collected annotation 
to a pre-prepared gold standard. We consider our outlier detection process 
an alternative as it keeps only those annotators who tend to agree with the others.}
For each evaluator we computed the distance of its average score from the average 
of the other evaluators and normalized by the standard deviation of the latter set. 
Evaluators whose statistic was above a predefined threshold \footnote{The threshold 
was set to 1.45, reasoning that if the statistics were sampled from a Gaussian 
with the empirical mean and variance, then $\sim$80\% of the evaluators would be included.} 
were excluded from the final dataset. 
We performed this procedure periodically and once a batch had 13 annotators that passed 
the test we stopped collecting scores for that batch.

\isection{Vector Space Models}{sec:vsm-models}


Here we describe the {\sc vsm}s we employ, their training data and evaluation protocol.

\subsection{Models}

\paragraph{Bag of Words ({\sc bow}).}
We constructed a {\sc vsm} following the optimal performance guidelines of \cite{kiela2014systematic}. 
After extracting the $k$ most frequent words in the training corpus, 
we generated a matrix of co-occurrence counts with a row for each of 
the words in any of the pairs in an evaluation set, and a column for each of the $k$ most frequent words.
Co-occurrence was counted within a window size $C$, without crossing sentence boundaries. 
The entries of the matrix were then normalized to PPMI values.
The resulting matrix's rows constitute the vector representations of the words.\footnote{We experimented with $k \in \{1000, 2000, \ldots, 10000\}$ and 
$C \in \{2, 3, \ldots, 8\}$ and set $k=10000$ and $C=2$ for all {\sc tl}s.} 

\paragraph{word2vec.}

The Mikolov et al.'s NN model 
\cite{Mikolov:2013a,Mikolov:2013c}.\footnote{
http://word2vec.googlecode.com/svn/trunk/word2vec.c}
The model aims to learn word representations that maximize the objective:

{\small \[L = \sum_{t=1}^{T}\sum_{-c \le j \le c, j\ne 0} \log p(w_{t+j}|w_t)\] }
Where $T$ is the number of training tokens, and $c$ a window size parameter. 
The objective respects sentence boundaries, conditioning only words from the same sentence on each other.
\footnote{We excluded this detail from the objective for brevity.}



We tuned three parameters $D$ - the vector dimensionality, 
$F$ - a frequency cutoff for words to be included in the objective, and $c$ - 
the window size. We followed Radim Rehurek's {\sc w2v} tutorial 
\footnote{http://radimrehurek.com/2014/02/word2vec-tutorial/} 
and set $c=5$, $D=400$ and $F=1$ for all {\sc tl}s.  

\subsection{Training and Word Pair Scoring}

We trained our {\sc vsm}s on the Wikipedia corpora released 
by \cite{al2014polyglot} .\footnote{https://sites.google.com/site/rmyeid/projects/polyglot} 
This is a set of multilingual \textit{comparable} corpora, as Wikipedia entries 
covering the same topic have similar content across languages. 
This allows us to focus on the effect of the {\sc (tl, jl)} combination, 
while keeping the training topics fixed across languages.

The size of these corpora is as follows (left number for the number of word types, 
right number for the number of word tokens): 
English (3.98 M, 1.4 G), German (5.1 M, 484.5 M), Italian (1.65 M, 281.6 M), 
Russian (2.81 M, 230 M).
Before training the models, we cleaned the corpora, removing stopwords and any string that is not comprised 
of alphabetic characters only,\footnote{ According to the NLTK list, http://www.nltk.org/} 
and stemming the remaining words using an NLTK stemmer.\footnote{http://www.nltk.org/howto/stem.html} 

The score assigned to a word pair by a model is the cosine similarity between 
the vectors the model induces for the pair's words. 
For each {\sc (tl, jl)} pair we compute the Spearman correlation coefficient 
($\rho$) between the ranking derived from a model's scores and the ranking derived from 
the human scores.\footnote{Result patterns 
are very similar when considering the Pearson and Kandall Tao scores. We hence keep our 
presentation concise and report only the Spearman scores.} 

Our main experimental setup reflects a preference for comparable corpora. 
This choice has consequences: first, our English and German corpora are substantially larger 
than their Russian and Italian counterparts; and, second, our training corpora are smaller 
than some of the alternative publicly available corpora that have been used for VSM training. 

To exclude the possibility that our observations are the mere outcome of these biases, 
we replicated the experiments of \secref {sec:training-language} and \secref {sec:training-language-combination} 
in two additional setups. First, in the \textit{small training setup} we re-ran our experiments when  
the English and the German training corpora were cut to the size of the Russian or the Italian corpus. 
The results in this setup were averaged over 5 random samples from each corpus. 
Second, in the \textit{large training setup} we re-ran our experiments when the English, German 
and Italian corpora were replaced with much larger, \textit {incomparable} corpora: 
English with the 8G word tokens corpus constructed using the {\sc w2v} script,
\footnote{code.google.com/p/word2vec/source/browse/trunk/demo-train-big-model-v1.sh} 
and Italian and German with the WaCky corpora (\cite{Baroni:2009},\footnote{http://wacky.sslmit.unibo.it/doku.php} 
Italian: 1.585G word tokens, German: 1.278G word tokens).\footnote{Russian is not included in this 
latter setup since we could not find a publicly available substantially larger Russian corpus.}
Since the result patterns in these setups are very similar to those in the major, 
comparable corpora setup, we report them briefly.


\isection{The Judgment Language Effect}{sec:judgment-language}

\begin{table*}[t!]
\centering
\scriptsize
\begin{tabular}{|l|l|l|l|l|l|l|l|l|}
\hline
 $L_1|L_2$ &\multicolumn{2}{|c|}{English} &\multicolumn{2}{|c|}{German} &\multicolumn{2}{|c|}{Italian} &\multicolumn{2}{|c|}{Russian} \\
\cline{1-9}
     & mean  & std & mean & std & mean & std & mean & std\\\hline
     
  English &  \textit {0.838} $|$  0.896 & \textit {0.083} $|$  0.033 & 0.752 & 0.105 & 0.739 & 0.092 & 0.739 & 0.110\\       
  \hline 
  German  & \textit {0.648} & \textit {0.187} & \textit {0.808} $|$  0.864 & \textit {0.062} $|$  0.055 & 0.700 & 0.105 & 0.720 & 0.076 \\  
   \hline     
  Italian   & \textit {0.729} & \textit {0.084} & \textit {0.633} & \textit {0.197} & \textit {0.879} $|$  0.871 &  \textit {0.053} $|$  0.055 & 0.720 & 0.121\\  
   \hline     
  Russian   & \textit {0.724} & \textit {0.097} & \textit {0.621} & \textit {0.170} & \textit {0.705} & \textit {0.073} & \textit {0.880} $|$  0.880 & \textit {0.045} $|$  0.033 \\  
   \hline
\end{tabular}
\caption{\footnotesize Average Spearman $\rho$ correlation coefficient between human judgments 
in the within and the cross language setups.  
The $(L_1,L_2)$ table entry (which is further divided into \textit{mean} and 
standard deviation (\textit{std}) columns) 
corresponds to the comparison of evaluators with judgment language $L_1$ to evaluators with judgment language $L_2$.
For each pair of languages the entry above the main diagonal of the matrix 
is for {\sc ws353} and the entry below the main diagonal (italic font) is for {\sc sl999} 
(for example, the (German, Italian) entry is for {\sc ws353} while the (Italian, German) 
entry is for {\sc sl999}). On the main diagonal, for both the mean and the std entries, 
the left number is for {\sc sl999}  while the right number is for {\sc ws353}.}
\label{table:annotator-correlation}
\end{table*}

\begin{table*}[t!]
\centering
\subfloat [{\sc bow} - {\sc ws353}]{
\scriptsize
\begin{tabular}{ |l|l|l|l|l|}
	\hline
        $T$ $|$ $J$ & English  & German & Italian & Russian\\ \hline
	English  & {\bf 0.600} & 0.523 & 0.488  & 0.496\\ \hline
	German   & 0.387       & {\bf 0.414} & 0.360  & 0.408 \\ \hline
	Italian  & {\bf 0.485} & 0.410 & 0.451  & 0.427\\ \hline
        Russian  & 0.403      & 0.377 & 0.360  & {\bf 0.426} \\ \hline
\end{tabular}
}
\quad
\subfloat [{\sc w2v} - {\sc ws353}]{
\scriptsize
\begin{tabular}{ |l|l|l|l|l|}
	\hline
        $T$ $|$ $J$ & English  &       German & Italian & Russian\\ \hline
English  & {\bf 0.652} & 	0.618 & 	0.614 & 	0.585   \\ \hline
German  & 0.537 & 	{\bf 0.595} & 	0.505 & 	0.554   \\ \hline
Italian &  0.564 & 	0.483 & 	{\bf 0.569} & 	0.504   \\ \hline
Russian  & 0.574 & 	0.532 & 	0.495 & 	{\bf 0.606}   \\ \hline
\end{tabular}
}
\quad
\subfloat [{\sc bow} - {\sc sl999}]{
\scriptsize
\begin{tabular}{ |l|l|l|l|l|}
	\hline
        $T$ $|$ $J$ & English  & German & Italian & Russian\\ \hline
	English  & 0.214       & {\bf 0.304} & 0.271  & 0.220\\ \hline
	German   & 0.086       & {\bf 0.268} & 0.199  & 0.087 \\ \hline
	Italian  & 0.140      &  {\bf 0.236} & 0.214  & 0.115\\ \hline
        Russian  & 0.141      &  {\bf 0.240} & 0.226  & 0.157 \\ \hline
\end{tabular}
}
\quad
\subfloat [{\sc w2v} - {\sc sl999}]{
\scriptsize
\begin{tabular}{ |l|l|l|l|l|}
	\hline
        $T$ $|$ $J$ & English  &       German & Italian & Russian\\ \hline
	English  & 0.266 & 	{\bf 0.354}  & 0.308  & 0.260\\ \hline
	German   & 0.198 & 	{\bf 0.342}  & 0.249  & 0.170\\ \hline
	Italian  & 0.207 &	{\bf 0.299}  & 0.293  & 0.197\\ \hline
        Russian  & 0.160 &	{\bf 0.250}	& 0.242  & 0.234\\ \hline
\end{tabular}
}
\caption{\footnotesize Spearman $\rho$ correlation coefficient between human scores 
and {\sc vsm} scores.
The $(T,J)$ entry of each matrix presents the $\rho$ value between the scores of a {\sc vsm} trained 
on language $T$ and the human scores produced for judgment language $J$. In each table, for each 
training language (row) the best judgment language is highlighted in bold.}
\label{table:train-judge-correlation}
\end{table*}

\begin{figure*}[t!]
  \centering
    \includegraphics[width=4cm,height=4cm]{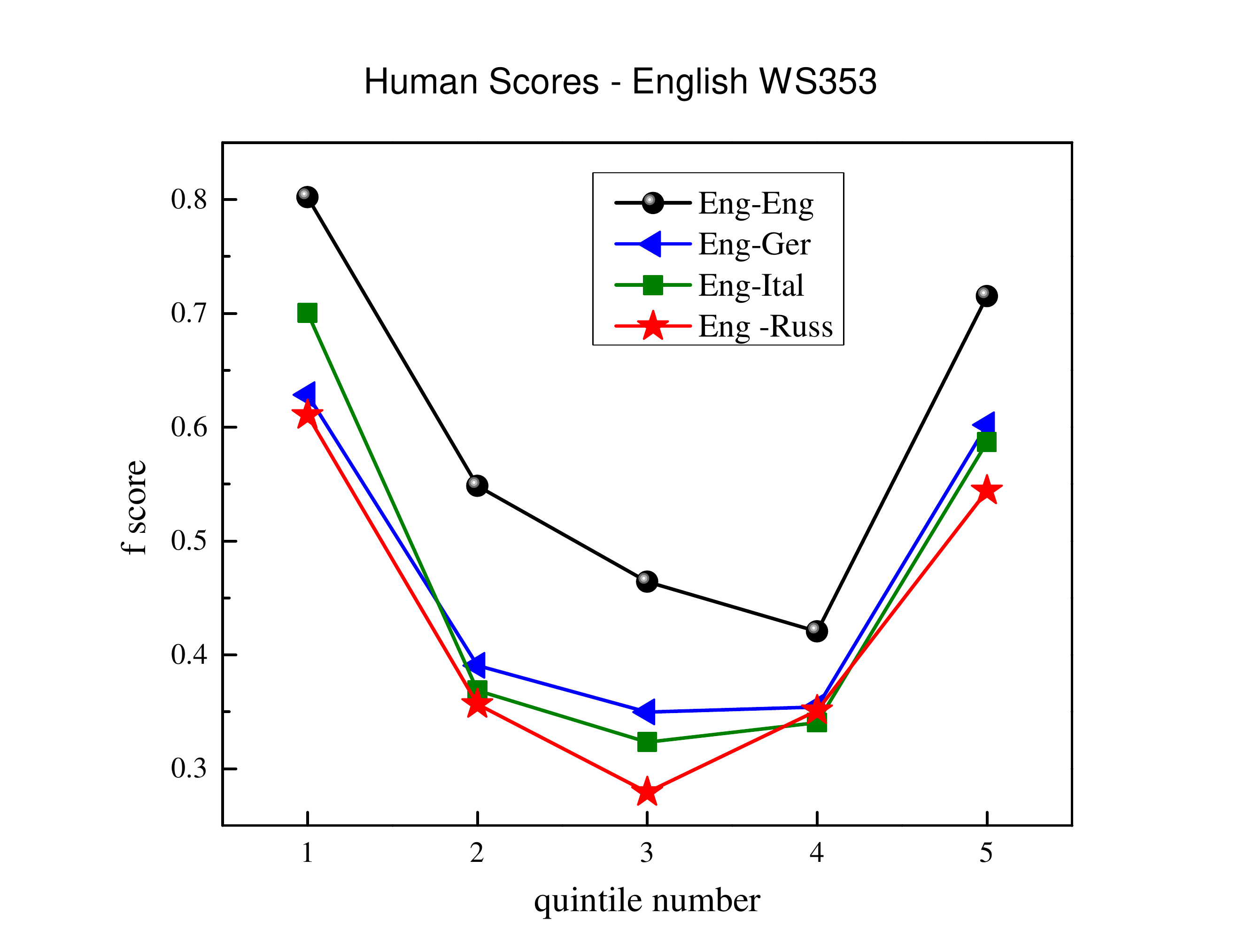}
    \includegraphics[width=4cm,height=4cm]{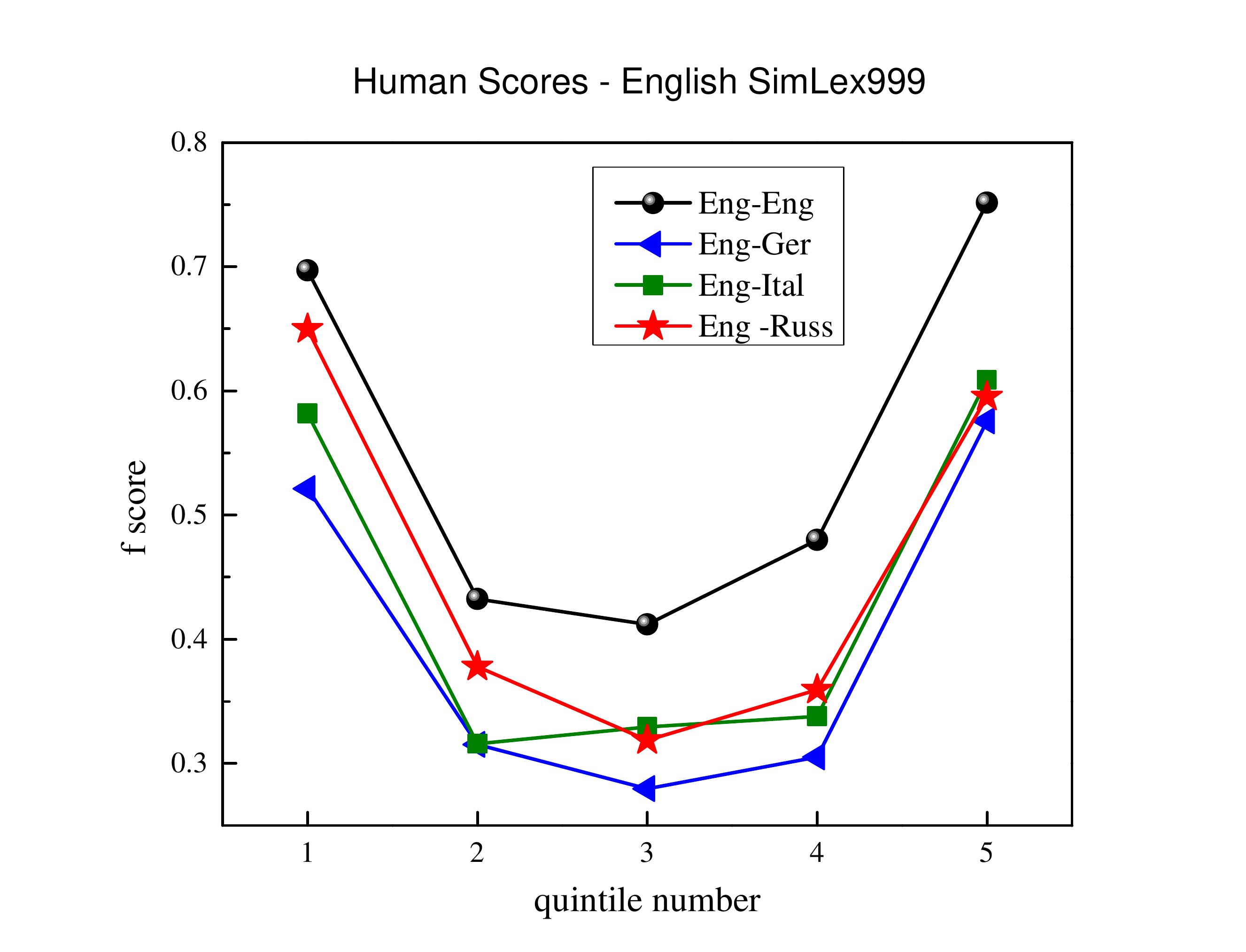}
    \includegraphics[width=4cm,height=4cm]{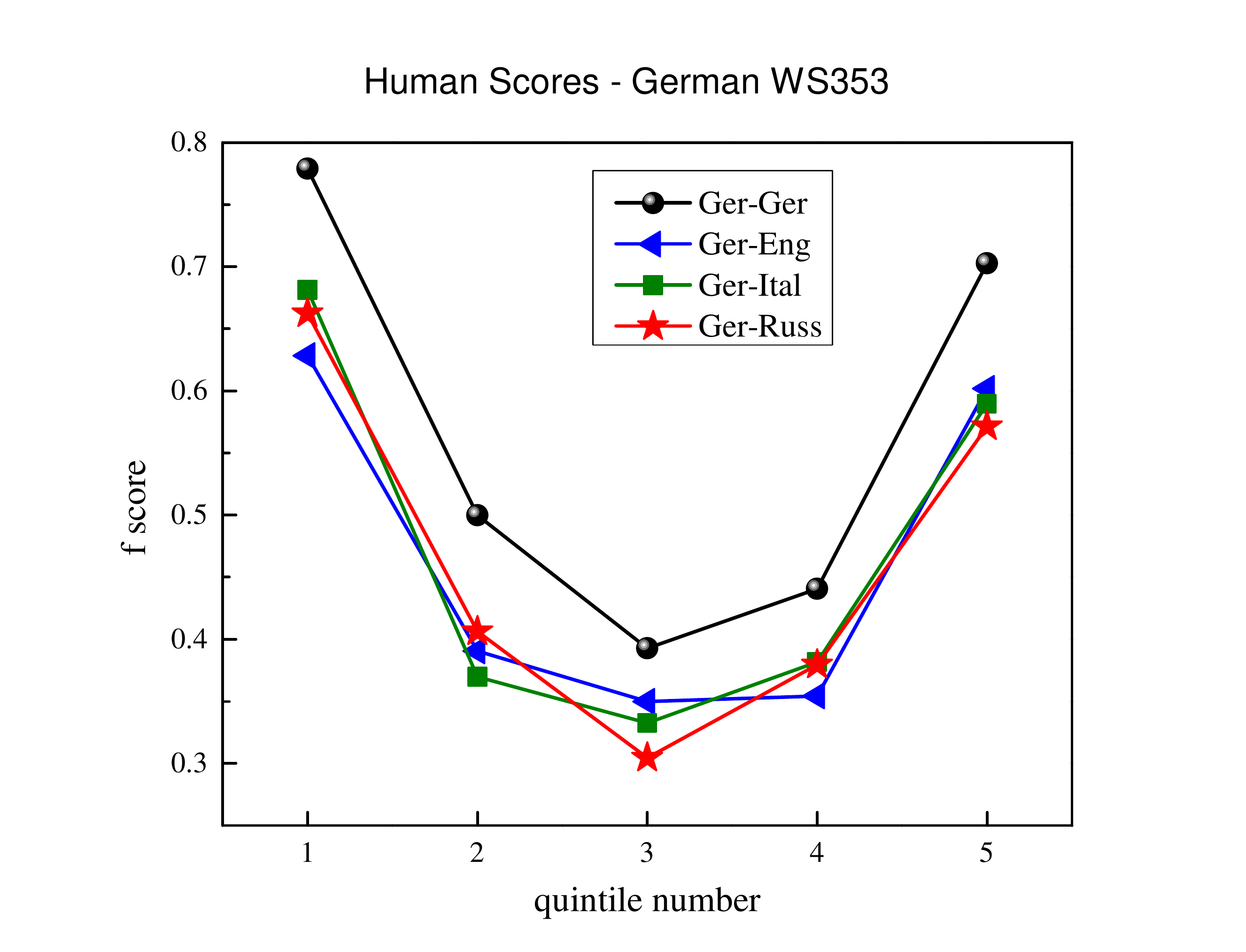}
    \includegraphics[width=4cm,height=4cm]{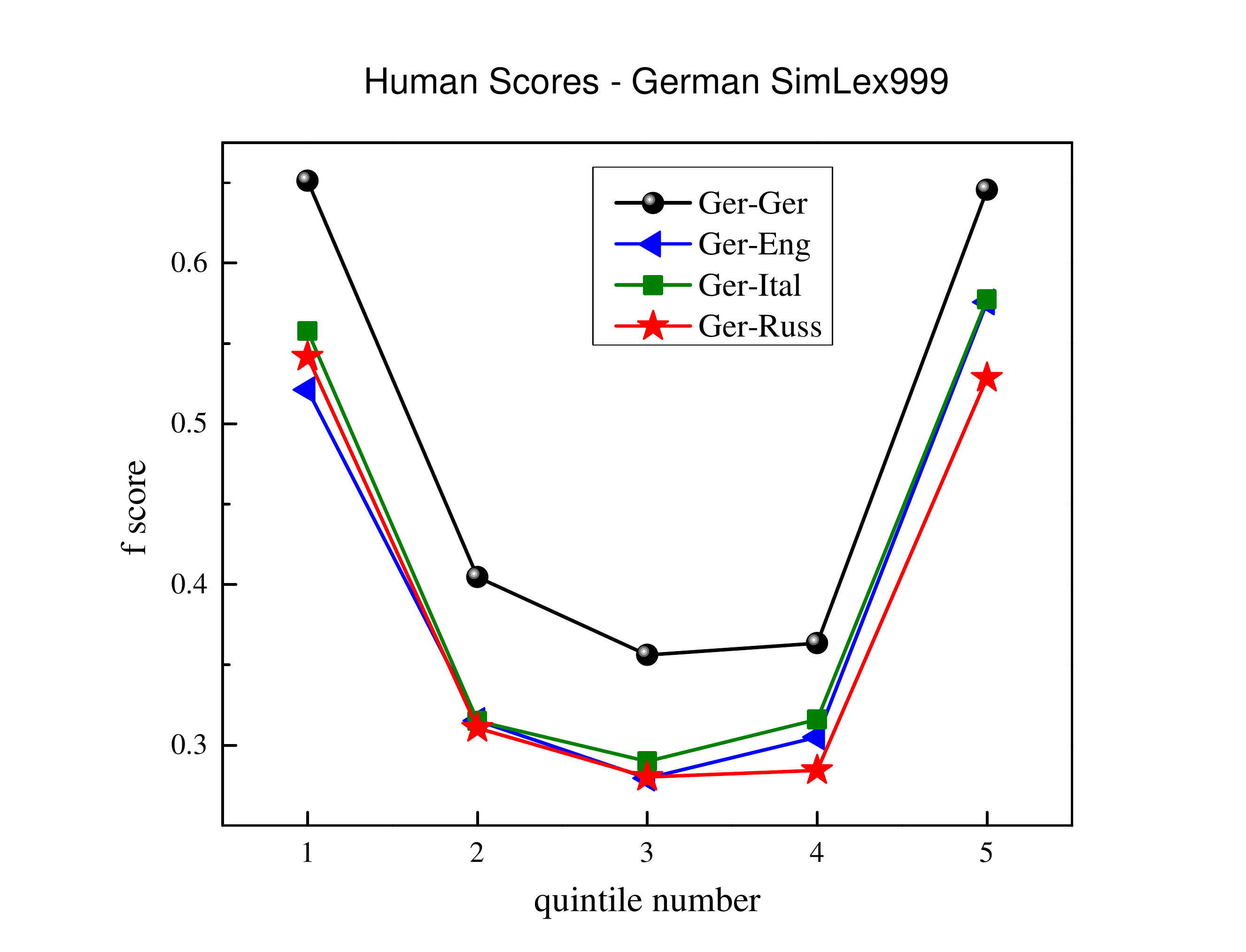}
    \includegraphics[width=4cm,height=4cm]{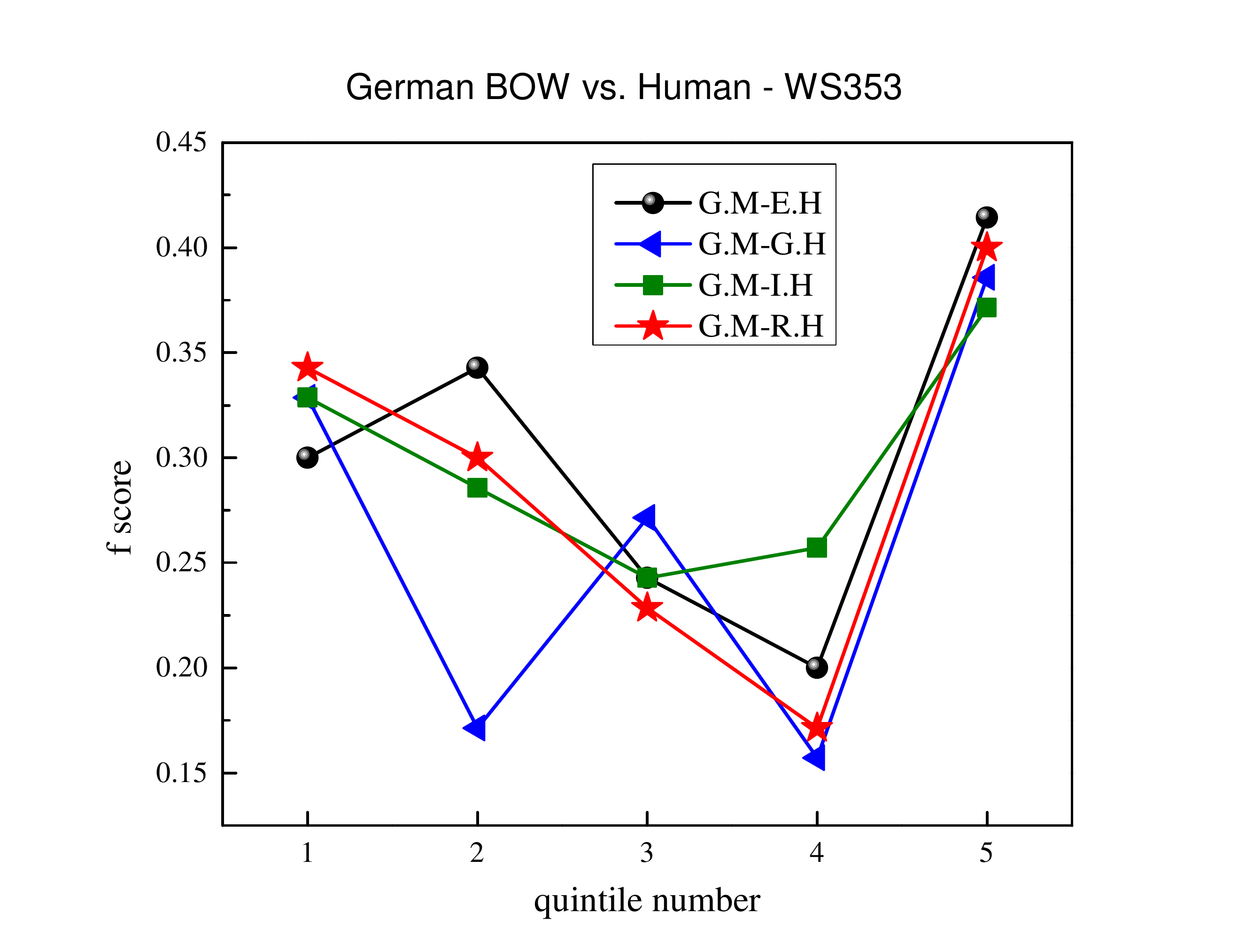}
    \includegraphics[width=4cm,height=4cm]{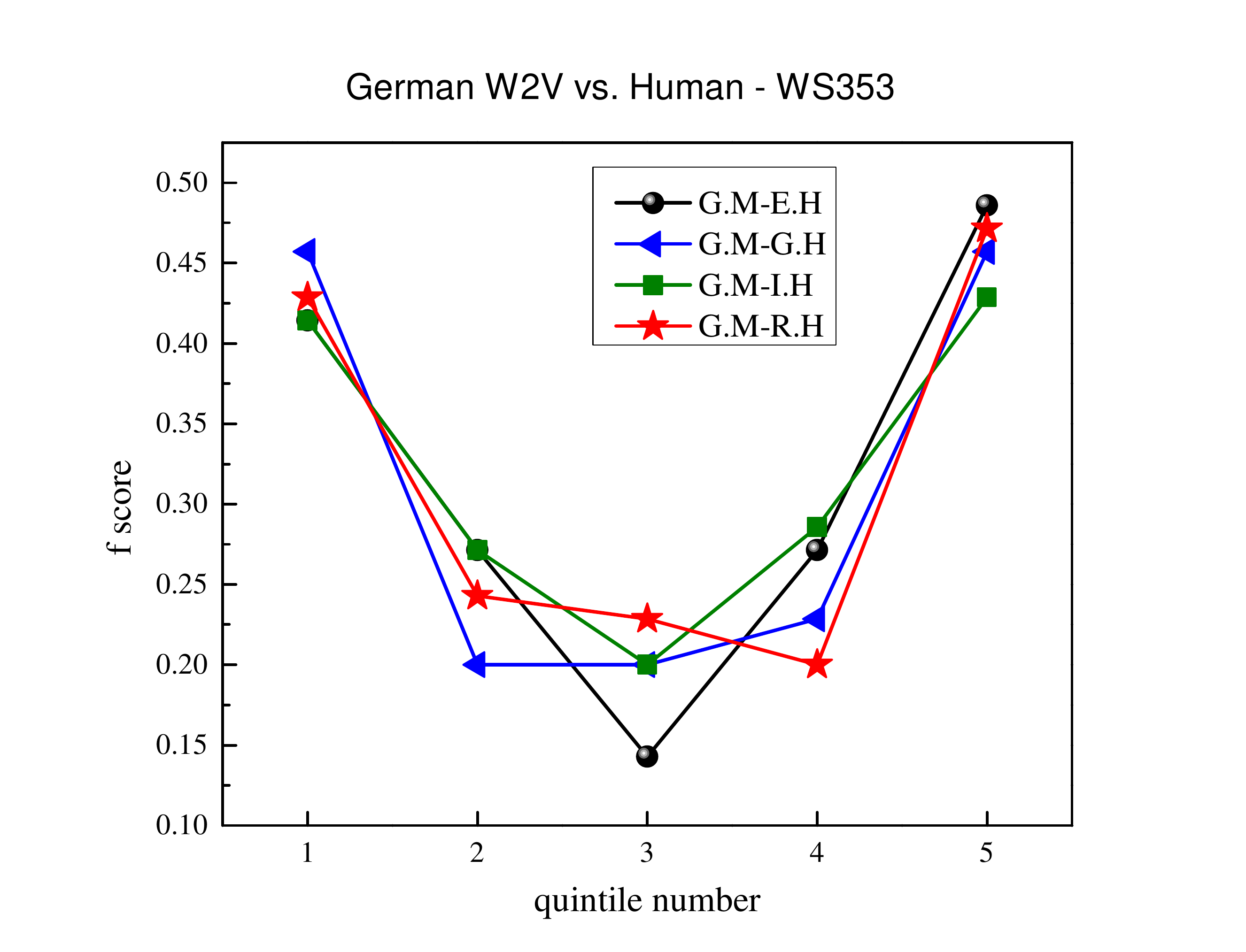}
    \includegraphics[width=4cm,height=4cm]{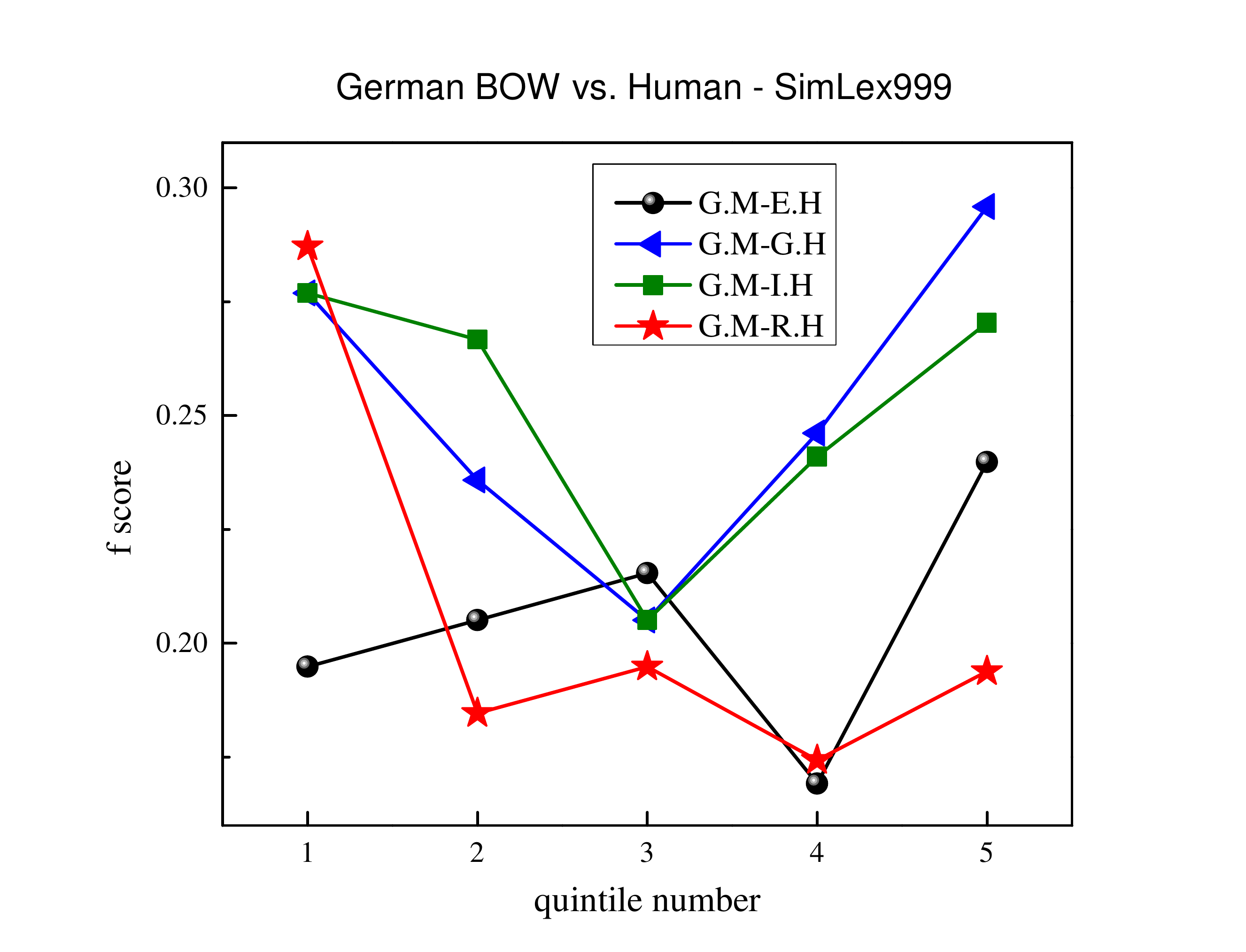}
    \includegraphics[width=4cm,height=4cm]{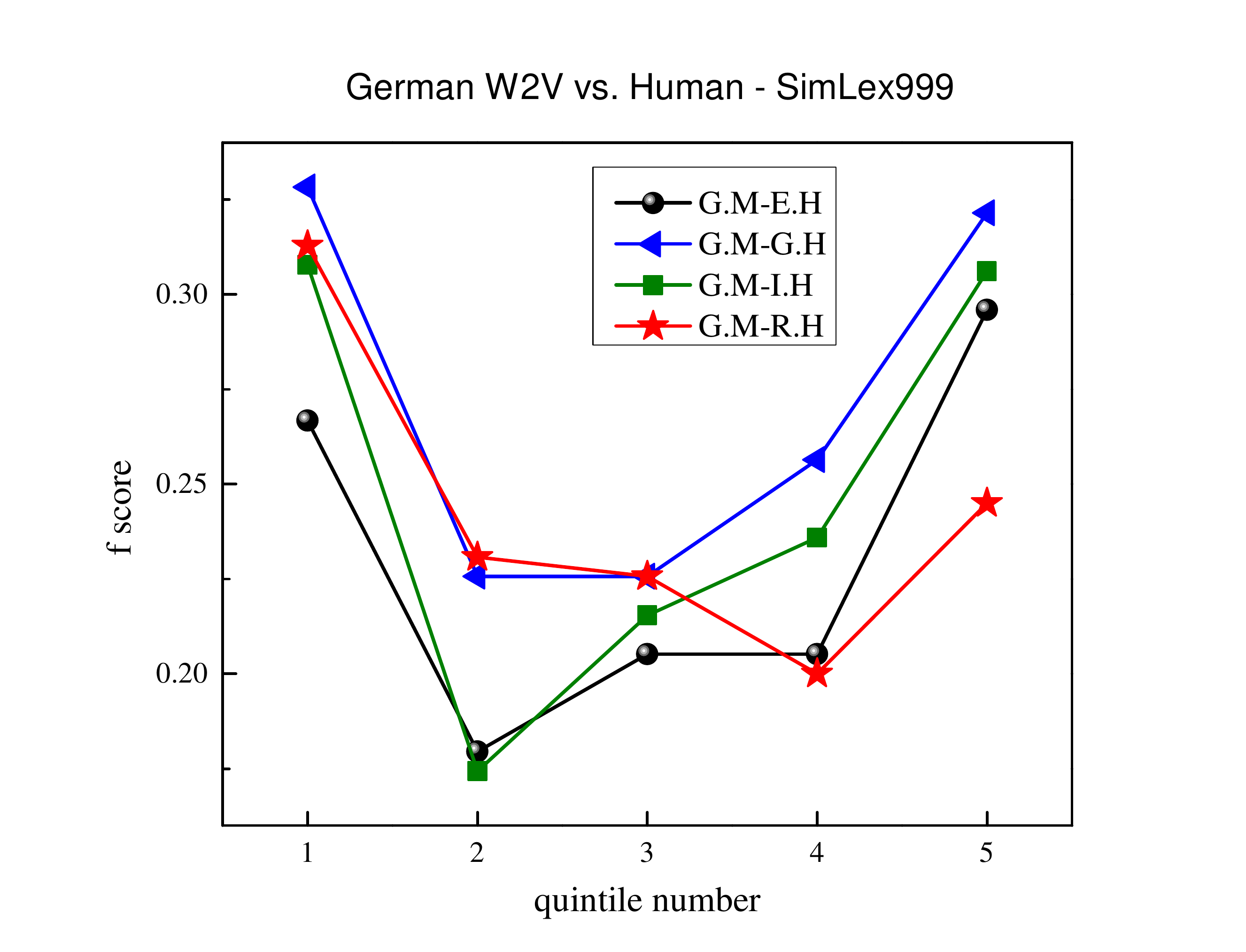}
  \caption{\footnotesize  Relative F-score of the word pair lists in corresponding 
quintiles of: (a) human rankings with different judgment languages (top line, two left graphs show all 
combinations of English with another language,  two right graphs show the same for German); and 
(b) model rankings with training language $l_1$ vs. human ranking with judgment 
language $l_2$ (bottom line, graphs presented for all combinations of models ({\sc bow} and {\sc w2v}) and evaluation sets ({\sc ws353} and {\sc sl999}) 
for $l1 = German$). Languages are denoted with a one or a three letter abbreviation, 
M stands for model and H for human.}
 \label{fig:ablation}
\end{figure*}


Our first question is: how does the {\sc jl} affect the word pair scores 
produced by the human evaluators. 
To provide a quantitative answer, we run the following protocol, both within and across {\sc jl}s.  
For each of the 50 word pair batches, we generate all possible 
$K$-size subsets of the batch evaluators, 
each $K$-size subset defining a unique partition of these evaluators (we set $K$ to 6). 
Then, for the within language evaluation we 
calculate the correlation between the averaged word pair scores  
of the two subsets induced by each K-size subset selection. 
For the cross language evaluation, in turn, we calculate 
the correlation between the average word pair scores of each K-size subset of language 1 
with its corresponding subset of language 2.
The resulting $\rho$ scores were averaged to get a final 
score for each language (in the within-language case) and language pair (in the cross-language case).\footnote {We 
have 1716 K-size subsets for each batch and totals of 1716*7 and 1716*20 correlations for each  
{\sc ws353} and {\sc sl999} scenarios respectively.}  

Table \ref{table:annotator-correlation} presents our results.
The correlations within a {\sc jl} are clearly higher compared to their cross {\sc jl} 
counterparts, with mean values at the range of $[0.864-0.896]$ for {\sc ws353} and $[0.808-0.880]$ 
for {\sc sl999} within {\sc jl}s, compared to $[0.700-0.752]$ for {\sc ws353} 
and $[0.621-0.729]$ for {\sc sl999} across {\sc jl}s. 

For both evaluation sets, we ran 
the Welch's t-test for each set of correlations computed for an individual language 
with each set of correlations computed for a pair of languages. 
In all 24 cases \footnote{we have four languages and hence six language pairs.} 
of each evaluation set the null hypothesis stating that the two sets have an 
equal mean was rejected with $Pvalue < 0.001$. 

Further, the standard deviation values are $[0.033-0.055]$ for {\sc ws353} and $[0.045-0.083]$ 
for {\sc sl999} in the within language setup, compared to $[0.076-0.121]$ for {\sc ws353} 
and $[0.073-0.197]$  for {\sc sl999} in the cross language setup. 
These results reflect the weaker dependence of the human judgment in the within 
language setup on the involved word-pairs and human evaluators.


To better understand the {\sc jl} effect, for each {\sc jl} we rank the word pairs 
according to their average human score and, then, compute for each pair of {\sc jl}s the relative 
F-score between corresponding quintiles in the rankings. \footnote{For each of the 1716 
K-size subset pairs (all possible divisions of the scores to subsets of 6 and 7 
for the within language case, every subset of size 6 in one language with its corresponding 
subset in the other language in the cross language case), 
we produced two word pair rankings according to the average score within each subset. 
We then divided each ranked list to 5 quintiles and computed relative F-scores 
between each pair of corresponding quintiles. 
We finally report the average F-score for each pair of corresponding quintiles across all these 1716 cases.} 
The top line of Figure \ref{fig:ablation} (two left graphs for comparisons where English is involved,
two right graphs for comparisons where German is involved) reveals 
that the overlap between corresponding quintiles is substantially larger for the top and bottom quintiles 
(top and bottom 20\% of the word pairs according to each of the rankings) compared to quintiles 2-4.
The graphs further demonstrate the larger overlap between corresponding quintiles in the 
within language setups compared to the cross-language setups, highlighting the impact 
of {\sc jl} differences on this phenomenon.\footnote{We performed the same analysis for the cases 
where the Italian and Russian {\sc jl}s are involved and observed very similar patterns. 
These graphs are omitted due to space constraints.}


All in all our results suggest that the concepts of word similarity and association may be {\sc jl} dependent.  
Our next natural question is how the relations between the {\sc vsm} {\sc tl} and the human {\sc jl} 
affect the correlation between the model and the human scores.
 
\isection{The VSM Training Language Effect}{sec:training-language}

Table \ref{table:train-judge-correlation} presents the 
Spearman $\rho$ correlation coefficient between human and model scores.

\paragraph{Training Language Choice.}
 
For each of the {\sc jl}s $J$, we first ask what is the {\sc tl} $T$ 
that leads to the monolingual model that best predicts human judgments with $J$. 
Both word association ({\sc ws353}) and similarity ({\sc sl999}) demonstrate very similar 
answers to this question.

A first shared pattern is that English is overall the best choice of {\sc tl} 
for both {\sc bow} and {\sc w2v}: in 7 out of 8 cases for {\sc ws353} and in all 8 cases for {\sc sl999}. 
A second shared pattern is that the {\sc jl} itself is overall the second best {\sc tl}, which 
is observed in 10 of the 11 cases where English is the best {\sc tl} 
for a given {\sc jl} and {\sc jl} != English. 

\paragraph{Judgment Language Choice.}

Our second question is complementary to the first one, namely, for each of the {\sc tl}s $T$, 
what is the {\sc jl} $J$ that leads to human judgments that best correlate 
with the predictions of the monolingual model trained with $T$. 
Here we observe considerable differences between  
word similarity and association. 

A first major difference is in the identity of the best {\sc jl}. 
While for {\sc ws353} in 7 of 8 cases a model trained with a {\sc tl} $T$ best correlates 
with human judgments made with $T$ as a {\sc jl}, 
for {\sc sl999} both models best correlate with German judgments for all {\sc tl}s.

A second major difference is related to the English {\sc jl}.
For {\sc ws353} in 3 out of the 5 cases where English is not the best {\sc jl} for a model it is 
the second best {\sc jl}. 
For {\sc sl999}, in contrast, for all 8 {\sc tl} and model type combinations, 
English is the {\sc jl} with the lowest correlation. For this dataset, Italian is always the second 
best {\sc jl}, and Russian is the third best.

\paragraph{VSM Comparison.}

Our experiments also cast light on the effectiveness of the participating {\sc vsm}s. 
For every combination of {\sc tl}, {\sc jl} 
and word pair dataset, the NN-based {\sc w2v} is superior to the count-based {\sc bow}. 
This finding supports recent 
conclusions on the superiority of "predict" models over their
"count" counterparts \cite{Baroni:2014}.

\paragraph {Quintile Analysis.}

To further investigate the mutual impact of the {\sc tl}s and {\sc jl}s, 
we replicated the quintile analysis of 
\secref{sec:judgment-language}, this time comparing the rankings of a model trained with language $l_1$ to 
the human scores obtained with {\sc jl} $l_2$.\footnote{Since the models produce only one score per word pair, 
in this analysis we ranked the word pairs according to the model scores as well as according to the average
of all 13 human scores, divided each ranked list to quintiles and computed a relative F-score 
for each pair of corresponding quintiles.} 
Results are presented in the bottom line 
of Figure \ref{fig:ablation}.\footnote{For brevity, we present only the curves for $l1 = German$, the 
patterns for the other cases are very similar.}

Interestingly, like in the respective analysis for {\sc jl} pairs, human-model 
disagreement is generally most prominent for word pairs that are considered of 
medium similarity or association. 
Note, however, that in the current analysis, the human-model agreement 
is weaker than the human-human agreement on the corresponding quintiles we explored in 
\secref{sec:judgment-language}. Moreover, while in the analysis of \secref{sec:judgment-language} 
the F-score values in the within 
language setup are superior to their cross-language counterparts, here keeping the 
{\sc tl} and {\sc jl} identical does not result in superior F-scores in most cases.

\begin{table*}[t!]
\centering
\subfloat [{\sc bow} - {\sc ws353}]{
\tiny
\begin{tabular}{ |l|l|l|l|l|l|}
	\hline                                                                          
        $T$ $|$ $J$ & English           &  German       &       Italian         &        Russian     \\ \hline        
E-G &                 0.544	& 	{\bf 0.528}	& 	{\bf 0.490}	& 	{\bf 0.504}  \\ \hline
E-I &                 0.575     &       {\bf 0.531}	& 	{\bf 0.516}	& 	{\bf 0.517}  \\ \hline
E-R &                 0.556	& 	{\bf 0.526}	& 	{\bf 0.494}	& 	{\bf 0.515}  \\ \hline
G-I &                 0.504	& 	{\bf 0.466}	& 	{\bf 0.479}	& 	{\bf 0.482}  \\ \hline
G-R &                 0.437	& 	{\bf 0.450}	& 	0.407  	        & 	{\bf 0.473}  \\ \hline
I-R &                 0.508	& 	{\bf 0.445}	& 	{\bf 0.475}	& 	{\bf 0.481}  \\ \hline \hline
E-E &                 0.543     & 	{\bf 0.518}     & 	{\bf 0.484}     & 	{\bf 0.492}   \\ \hline
G-G &                 0.342     &  	0.400	        &       0.353           & 	0.402         \\ \hline
I-I &                 0.46      & 	0.400	        &       0.443           & 	0.416         \\ \hline
R-R &                 0.395     & 	0.365           & 	0.355           & 	0.420          \\ \hline
\end{tabular}
}
\quad
\subfloat [{\sc w2v} - {\sc ws353}]{
\tiny
\begin{tabular}{ |l|l|l|l|l|}
	\hline            
        $T$ $|$ $J$ &   English         &          German       &       Italian         &        Russian  \\ \hline        
E-G &                   {\bf 0.675}	& 	{\bf 0.667}	& 	{\bf 0.630}	& 	{\bf 0.630} \\ \hline
E-I &                   {\bf 0.696}	& 	{\bf 0.633}	& 	{\bf 0.662}	& 	{\bf 0.621} \\ \hline
E-R &                   {\bf 0.687}	& 	{\bf 0.645}	& 	{\bf 0.624}	& 	{\bf 0.646} \\ \hline
G-I &                   {\bf 0.657}	& 	{\bf 0.629}	& 	{\bf 0.625}	& 	{\bf 0.623} \\ \hline
G-R &                   0.618	        & 	{\bf 0.628}	& 	0.553   	& 	{\bf 0.645} \\ \hline
I-R &                   {\bf 0.657}	& 	0.585	        &       {\bf 0.606}	& 	{\bf 0.644} \\ \hline \hline
E-E &                   0.652	        &       {\bf 0.620}     & 	{\bf 0.609}     & 	0.591       \\ \hline
G-G &                   0.540           & 	{\bf 0.601}     & 	0.497           & 	0.563       \\ \hline
I-I &                   0.582           & 	0.500	        &       {\bf 0.582}     & 	0.514       \\ \hline
R-R &                   0.571           & 	0.534           & 	0.498           & 	{\bf 0.610}  \\ \hline
\end{tabular}
}
\quad
\subfloat [{\sc bow} - {\sc sl999}]{
\tiny
\begin{tabular}{ |l|l|l|l|l|}
	\hline         
        $T$ $|$ $J$ & English   &        German         &        Italian        &        Russian  \\ \hline        
E-G &                0.177	& 	{\bf 0.334}	& 	{\bf 0.273}	& 	{\bf 0.180}  \\ \hline
E-I &                0.201	& 	{\bf 0.313}	& 	{\bf 0.276}	& 	{\bf 0.192}  \\ \hline
E-R &                0.209	& 	{\bf 0.318}	& 	{\bf 0.289}	& 	{\bf 0.217}  \\ \hline
G-I &                0.131	& 	{\bf 0.294}	& 	{\bf 0.238}	& 	0.119  \\ \hline
G-R &                0.135	& 	{\bf 0.288}	& 	{\bf 0.243}	& 	0.145  \\ \hline
I-R &                0.164	& 	{\bf 0.281}	& 	{\bf 0.256}	& 	{\bf 0.164}  \\ \hline \hline
E-E &                0.210      & 	{\bf 0.302}     & 	{\bf 0.267}     & 	{\bf 0.215}   \\ \hline
G-G &                0.078      &  	0.259           & 	0.194           & 	0.078        \\ \hline
I-I &                0.137	&       0.235           & 	0.209           & 	0.110          \\ \hline
R-R &                0.137	&       0.235           & 	{\bf 0.223}     & 	0.150         \\ \hline

\end{tabular}
}
\quad
\subfloat [{\sc w2v} - {\sc sl999}]{
\tiny
\begin{tabular}{ |l|l|l|l|l|}
	\hline                       
        $T$ $|$ $J$               & English     &       German          &        Italian        &        Russian  \\ \hline        
E-G &                             0.263	        & 	{\bf 0.392}	& 	{\bf 0.313}	& 	{\bf 0.244}  \\ \hline
E-I &                            {\bf 0.267}	& 	{\bf 0.371}	& 	{\bf 0.340}	& 	{\bf 0.260}  \\ \hline
E-R &                             0.233	        & 	0.332    	& 	{\bf 0.302}	& 	{\bf 0.274}  \\ \hline
G-I &                             0.242  	& 	{\bf 0.380}	& 	{\bf 0.319}	& 	    0.223    \\ \hline
G-R &                             0.212  	& 	0.338	        & 	0.284    	& 	{\bf 0.242}  \\ \hline
I-R &                             0.207	        & 	0.311	        & 	{\bf 0.303}	& 	{\bf 0.250}  \\ \hline  \hline
E-E &                             0.261         & 	{\bf 0.353}     & 	{\bf 0.311}     & 	{\bf 0.254}  \\ \hline
G-G &                             0.196         & 	{\bf 0.344}     & 	0.248           & 	0.169         \\ \hline
I-I &                             0.206         & 	0.307           & 	{\bf 0.300}     & 	0.195         \\ \hline
R-R &                             0.170        & 	0.256           & 	0.260           & 	{\bf 0.241}    \\ \hline
\end{tabular}
}
\caption{\footnotesize Spearman $\rho$ correlation coefficient between human scores and the outcome of 
a linear interpolation ({\sc li}) of the scores of pairs of monolingual models. 
The $(T = L_1-L2,J = L3)$ entry of each table is the correlation of 
(1) the outcome of a {\sc li} of the scores of monolingual models 
trained on languages $L_1$ and $L_2$ with (2) the human scores 
produced with the {\sc jl} $L_3$. Cases where the {\sc li} of $L1$ and $L_2$ outperforms 
a monolingual  model trained on $L_3$ (where $L_3$ is the {\sc jl}) are highlighted in bold.
.}
\label{table:language-interpolation}
\end{table*}

\begin{table*}[t!]
\centering
\subfloat [{\sc bow} - {\sc ws353}]{
\tiny
\begin{tabular}{ |l|l|l|l|l|}
	\hline
$T$ $|$ $J$ &    English  &  German      & Italian & Russian  \\ \hline
E-G &    -0.130	 &         -0.071 & 	-0.070	 & -0.121  \\ \hline
E-I &    -0.068 & 	-0.033 & 	-0.011 & 	-0.070  \\ \hline
E-R &    -0.099	 & -0.043 & 	-0.026 & 	-0.090  \\ \hline
G-I &  -0.076 & 	-0.052 & 	-0.038	 & -0.081  \\ \hline
G-R &   -0.128	 & -0.068 & 	-0.075 & 	-0.126  \\ \hline
I-R &   -0.062 & 	-0.034 & 	-0.032	 & -0.084  \\ \hline \hline
E-E &    -0.112 & 	-0.049 & 	-0.034 & 	-0.103 \\ \hline
G-G &    -0.14  & 	-0.078 & 	-0.107 & 	-0.129 \\ \hline
I-I &    -0.06 & 	-0.036 & 	-0.006 & 	-0.059 \\ \hline
R-R &    -0.112 & 	-0.059 & 	-0.039 & 	-0.103 \\ \hline
\end{tabular}
}
\quad
\subfloat [{\sc w2v} - {\sc ws353} ]{
\tiny
\begin{tabular}{ |l|l|l|l|l|}
	\hline
        $T$ $|$ $J$ & English  &  German      & Italian & Russian  \\ \hline
E-G &      0.340 & 	0.356 & 	0.312 & 	0.273  \\ \hline
E-I &      0.295 & 	0.289 & 	0.302 & 	0.259  \\ \hline
E-R &      0.274 & 	0.291 & 	0.261 & 	0.286  \\ \hline
G-I &      0.286 & 	0.311 & 	0.291 & 	0.268  \\ \hline
G-R &      0.284 & 	0.339 & 	0.267 & 	0.307  \\ \hline
I-R &      0.223 & 	0.218 & 	0.230 & 	0.261  \\ \hline \hline
E-E &      0.28	 &  0.251 & 	0.23 & 	    0.226 \\ \hline
G-G &      0.206 & 	0.291 & 	0.212 & 	0.219 \\ \hline
I-I &      0.228 & 	0.227 & 	0.205 & 	0.147 \\ \hline
R-R &      0.236 & 	0.252 & 	0.253 & 	0.297 \\ \hline
\end{tabular}
}
\quad
\subfloat [{\sc bow} - {\sc sl999}]{
\tiny
\begin{tabular}{ |l|l|l|l|l|}
\hline                    
$T$ $|$ $J$ & English  &  German      & Italian     &  Russian  \\ \hline
E-G & {\bf 0.222}  &    0.234 & 	{\bf 0.273} & 	{\bf 0.253}   \\ \hline
E-I & {\bf 0.232}  & 	0.214 & 	{\bf 0.260} & 	{\bf 0.236}   \\ \hline
E-R & {\bf 0.270}  & 	0.240 & 	{\bf 0.289} & 	{\bf 0.277}   \\ \hline
G-I & 0.199        & 	0.211 & 	{\bf 0.249} & 	{\bf 0.216}   \\ \hline
G-R & 0.212         &    0.206 & 	{\bf 0.246} & 	{\bf 0.228}   \\ \hline
I-R & {\bf 0.223}   &    0.192 & 	{\bf 0.239} & 	{\bf 0.226}   \\ \hline \hline
E-E &  {\bf 0.244} & 	0.226 & 	{\bf 0.274} & 	{\bf 0.258} \\ \hline
G-G &  0.149 & 	0.185 & 	{\bf 0.215}               & 	{\bf 0.182} \\ \hline
I-I &  0.183 & 	0.175 & 	0.209               & 	{\bf 0.176} \\ \hline
R-R &  0.191 & 	0.189 & 	{\bf 0.221}         & 	{\bf 0.198} \\ \hline
\end{tabular}
}
\quad
\subfloat [{\sc w2v} - {\sc sl999}]{
\tiny
\begin{tabular}{ |l|l|l|l|l|}
\hline         
$T$ $|$ $J$ & English   &  German      & Italian & Russian  \\ \hline
E-G & {\bf 0.319}       & {\bf 0.434}   & {\bf 0.374} &	{\bf 0.296}  \\ \hline
E-I & {\bf 0.312}	& {\bf 0.408}	& {\bf 0.395} &	{\bf 0.297}   \\ \hline
E-R & {\bf 0.296}	& {\bf 0.361}   & {\bf 0.355} &	{\bf 0.320}   \\ \hline
G-I & {\bf 0.289}	& {\bf 0.417}	& {\bf 0.369} &	{\bf 0.257}    \\ \hline
G-R & {\bf 0.275}	& {\bf 0.366}	& {\bf 0.330} &	{\bf 0.287}   \\ \hline
I-R & 0.262	        & 0.331	        & {\bf 0.346} &	{\bf 0.287}    \\ \hline \hline
E-E & {\bf 0.332}       & {\bf 0.399}   & {\bf 0.380}  & {\bf 0.310}  \\ \hline
G-G & 0.251             & {\bf 0.394}   & {\bf 0.310}  & {\bf 0.237} \\ \hline
I-I & 0.250            & {\bf 0.353}   & {\bf 0.358} & 	0.217 \\ \hline
R-R & 0.240             & 	0.294   & {\bf 0.309} & {\bf 0.288} \\ \hline
\end{tabular}
}
\caption{\footnotesize Spearman $\rho$ correlation coefficient of the scores resulting 
from a {\sc cca} combination of monolingual models, with corresponding human scores. 
Table entries and highlighting is as in Table \ref{table:language-interpolation}.
}
\label{table:CCA}
\end{table*}

\paragraph {Observations.}

Our analysis leads to several observations. 
First, word similarity and association judgments have a language 
specific component. Consequently, the {\sc jl} is a good choice for model 
training (first question) and the predictions of models trained on a given language 
are best correlated with human judgments performed with that language, at 
least for word association (second question).
While this seems obvious in machine learning terms, as in-domain training is 
preferable and language change is analogous to domain change, 
the semantic nature of our tasks would suggest 
that {\sc vsm}s should preserve their outcome across languages. Our results suggest that this
latter assumption is not true.

Second, English has a special status in {\sc vsm} research: as a {\sc vsm} {\sc tl} for 
both association and similarity prediction (first question), and as a {\sc jl} 
for word association. 
The special status of English as a {\sc tl} may result from its simpler 
morphology \footnote{This is reflected, for example, by the lower type-to-token ratio of English in our 
training corpora: English = 0.0028; German = 0.011; Italian = 0.0058; Russian = 0.012.} 
which may allow more robust statistics to be collected. Another possible explanation is that
our evaluators are likely to have some command of English \footnote{We have not checked this.} which 
may bias their semantic judgments towards those made by an English trained model.

The {\sc jl} pattern is harder to understand. One possible hypothesis is that 
the dominance of English for word association is the result of our evaluation 
sets being translations of sets originally authored in English. Consequently, some 
important meaning components may get lost in translation. 
However, the poor similarity predictions of both models with all four {\sc tl}s
when English is the {\sc jl}, seriously challenge this hypothesis.

Finally, for word similarity both {\sc vsm}s are much better correlated with 
human scores when the {\sc jl} is German compared to the other {\sc jl}s and particularly to English. 
We will investigate this surprising observation in future work.


\paragraph {Training Corpus Size Effect.}

In the \textit {small training setup} our results were very similar to the results reported above 
both in terms of qualitative patterns and in the numerical correlation values 
(up to 0.02 difference in Spearman $\rho$). 
In the \textit {large training setup} we observed the exact same patterns detailed above 
but the Spearman $\rho$ values for every {\sc (tl, jl)} pair were higher than those 
of Table \ref{table:train-judge-correlation}, with the $\rho$ differences having the following (mean, std) values:
{\sc bow/ws353}: (0.036,0.025), {\sc w2v/ws353}: (0.031,0.013), 
{\sc bow/sl999}: (0.048,0.046) and {\sc w2v/sl999}: (0.042,0.033).

Our final investigation is of the potential of monolingual {\sc vsm} combination to compensate for the 
{\sc jl} effect. 

\isection{The Multilingual Combination Effect}{sec:training-language-combination}

We explore two simple methods for the combination of 
{\sc vsm}s trained on corpora of different languages, $l1$ and $l2$. 
In the first method, \textit{linear interpolation ({\sc li})}, we combine the 
scores produced by two {\sc vsm}s for a word pair $(w_i,w_j)$ using the linear equation: 

{\small \[ Score (w_i,w_j) = \lambda \cdot sc_{l1}(w_i,w_j) + (1 - \lambda) \cdot sc_{l2}(w_i,w_j)\]}
Where $sc_{li}(w_i,w_j)$ is the score produced by the model trained on the $li$ language 
and $\lambda \in [0,1]$.\footnote{
We experimented with $\lambda \in \{0.25,0.33,0.5,0.67,0.75\}$
and got improvements for most combinations of {\sc tl} pairs, 
{\sc jl}s and $\lambda$s (see below). 
We report results with $\lambda = 0.5$, giving both monolingual models an equal weight.} 

Our second combination method is Canonical Correlation Analysis ({\sc cca}). 
For each pair of languages, $(l_1,l_2)$, we calculated a pair 
of projection matrices to the shared subspace through the {\sc cca} method \cite{hardoon2004canonical}, 
using the vectors induced by monolingual models trained on an $l_1$  and an $l_2$ corpora. 
We then constructed a multilingual vector representation for each word by 
concatenating the $l_1$ and $l_2$ projected representations.\my{Should we give more details on CCA like some 
reviewers, and particularly, EMNLP rev1 asked for ?}
\footnote {Following \cite{faruqui2014improving} we also experimented with  
taking one of the monolingual projected vectors as the multilingual 
representation and got very similar results.}
\footnote{We applied both protocols for the combination of three and four monolingual models  
and did not observe substantial improvements over two-language multilingual models.}


We compare the performance of each multilingual combination method to a monolingual baseline in which 
the predictions of two monolingual models, each trained with randomly sampled 80\% of {\it the same} 
monolingual training corpus, were combined using one of the above methods.\footnote{These results were 
averaged over 5 random samples from each of the corpora. For {\sc li} we naturally 
used $\lambda = 0.5$. For {\sc cca} we employed the same protocol as in multilingual combination.}
This is done in order to rule out the possibility that our improvements are the mere result of the 
smoothing effect that model combination provides.

Tables \ref{table:language-interpolation} (top 6 lines of each table) presents 
results for multilingual {\sc li}.  
The numbers clearly show that this is an effective method of combining two 
monolingual models, leading to improvements over the participating monolingual models 
in most dataset and model combinations.\footnote{This effect is not highlighted in the table but is evident
from a comparison  to the numbers reported in Table \ref{table:train-judge-correlation}.}
Improvements computed with respect to monolingual models trained on the {\sc jl}
({\sc tl} = {\sc jl}, i.e. the results on the main diagonals of the sub-tables of table~\ref{table:train-judge-correlation}), 
are more prominent for German, Italian and Russian than for English (highlighted in bold in Table 
\ref{table:language-interpolation}), which is not surprising 
given that English is the best {\sc tl} of monolingual {\sc vsm}s for the majority of evaluation 
set, {\sc jl} and model combinations (\secref{sec:training-language}). 
Multilingual interpolated models improve over such non-interpolated monolingual models 
in 68 of 96 cases (70.8\%). 

Comparison to monolingual {\sc li} (bottom 4 lines of each table) 
reveals the impact of the multilingual combination. As an example indication,
monolingual {\sc li} improves over monolingual models trained on the {\sc jl} in only 
18 of 64 cases (28.1 \%).\footnote{A simple concatenation of 
the monolingual vectors is also an effective combination method of monolingual models, 
leading to improvements that are similar to what we report for {\sc li}. However, simple concatenation 
is effective for the {\sc bow} model only when PPMI normalization is applied to the row counts, 
as opposed to {\sc li} which is effective regardless of this step. 
We therefore focus on {\sc li}, the more robust method.}


Interestingly, {\sc cca} combination improves over monolingual models 
trained on the {\sc jl} only for {\sc sl999}. 
This result adds mixed observations to previous positive results on the effect of {\sc cca} combination 
for multilingual {\sc vsm} construction with the English {\sc jl} \cite{faruqui2014improving} 
and for the combination of visual and textual representations \cite{silberer2012grounded,hill2014multi}.

Like in \secref{sec:training-language}, we controlled against corpus size effects.
The results of both the \textit{small} and the \textit{large training setups} 
were very similar to those reported above both qualitatively and quantitatively. For example, in the \textit{large training setups}
the multilingual interpolated models improved over monolingual non-interpolated models trained
on the {\sc jl} in 61.1\% of the cases, compared to 16.7\% of the cases where the monolingual interpolated
models achieve such an improvement. The differences in numerical Spearman $\rho$ values were up to 0.01 
across setups.

\isection{Conclusions}{sec:discussion}

In this paper we aimed to establish the importance of the human {\sc jl} in lexical semantics research. 
We translated and re-scored two prominent datasets, {\sc ws353} and {\sc sl999}, and demonstrated the impact of 
the {\sc jl} on: (a) human semantic judgments; and (b) the correlation of monolingual and multilingual 
VSM predictions, produced with varios training languages, with human judgments. 


In future work we intend to extend our inquiry to relations beyond word association and similarity 
and to a larger number of {\sc tl}s and {\sc jl}s. We further intend to 
explore more advanced methods for multilingual {\sc vsm} construction. 
Finally, we would like to go beyond quantitative analysis and 
identify qualitative patterns in our data. 
Our ultimate goal is to construct {\sc vsm}s that directly account for 
the relations between their {\sc tl}(s) and potential human {\sc jl}s.

\bibliographystyle{acl2012}

\bibliography{cross-lingual-VSMs}

\end{document}